\documentclass[sigconf]{acmart}

\usepackage{algorithm}
\usepackage{algorithmic}
\usepackage{pifont}

\AtBeginDocument{%
  }

\copyrightyear{2026}
\acmYear{2026}
\setcopyright{cc}
\setcctype{by}
\acmConference[ICMR '26]{International Conference on Multimedia Retrieval}{June 16--19, 2026}{Amsterdam, Netherlands}
\acmBooktitle{International Conference on Multimedia Retrieval (ICMR '26), June 16--19, 2026, Amsterdam, Netherlands}
\acmDOI{10.1145/3805622.3810860}
\acmISBN{979-8-4007-2617-0/2026/06}

\begin{document}

\title{EAD-Net: Emotion-Aware Talking Head Generation with Spatial Refinement and Temporal Coherence}

\author{Yahui Li}
\email{lyh12092022@163.com}
\affiliation{%
  \institution{School of Computer Science and Technology, Xinjiang University}
  \city{Urumqi}
  \country{China}
}

\author{Yinfeng Yu}
\email{yuyinfeng@xju.edu.cn}
\affiliation{%
  \institution{School of Computer Science and Technology, Xinjiang University}
  \city{Urumqi}
  \country{China}
}

\author{Liejun Wang}
\authornote{Corresponding author}
\email{wljxju@xju.edu.cn}
\affiliation{%
  \institution{School of Computer Science and Technology, Xinjiang University}
  \city{Urumqi}
  \country{China}
}

\author{Shengjie Shen}
\email{shenshengjie1122@163.com}
\affiliation{%
  \institution{School of Computer Science and Technology, Xinjiang University}
  \city{Urumqi}
  \country{China}
}

\renewcommand{\shortauthors}{Li et al.}

\begin{abstract}
Emotional talking head video generation aims to generate expressive portrait videos with accurate lip synchronization and emotional facial expressions. Current methods rely on simple emotional labels, leading to insufficient semantic information. While introducing high-level semantics enhances expressiveness, it easily causes lip-sync degradation. Furthermore, mainstream generation methods struggle to balance computational efficiency and global motion awareness in long videos, and suffer from poor temporal coherence. Therefore, we propose an \textbf{E}motion-\textbf{A}ware \textbf{D}iffusion model-based \textbf{Net}work, called \textbf{EAD-Net}. We introduce SyncNet supervision and Temporal Representation Alignment (TREPA) to mitigate lip-sync degradation caused by multi-modal fusion. To model complex spatio-temporal dependencies in long video sequences, we propose a Spatio-Temporal Directional Attention (STDA) mechanism that captures global motion patterns through strip attention. Additionally, we design a Temporal Frame graph Reasoning Module (TFRM) to explicitly model temporal coherence between video frames through graph structure learning. To enhance emotional semantic control, a large language model is employed to extract textual descriptions from real videos, serving as high-level semantic guidance. Experiments on the HDTF and MEAD datasets demonstrate that our method outperforms existing methods in terms of lip-sync accuracy, temporal consistency and emotional accuracy.
\end{abstract}



\maketitle

\begin{figure}[t]
    \centering
    \includegraphics[width=1\columnwidth]{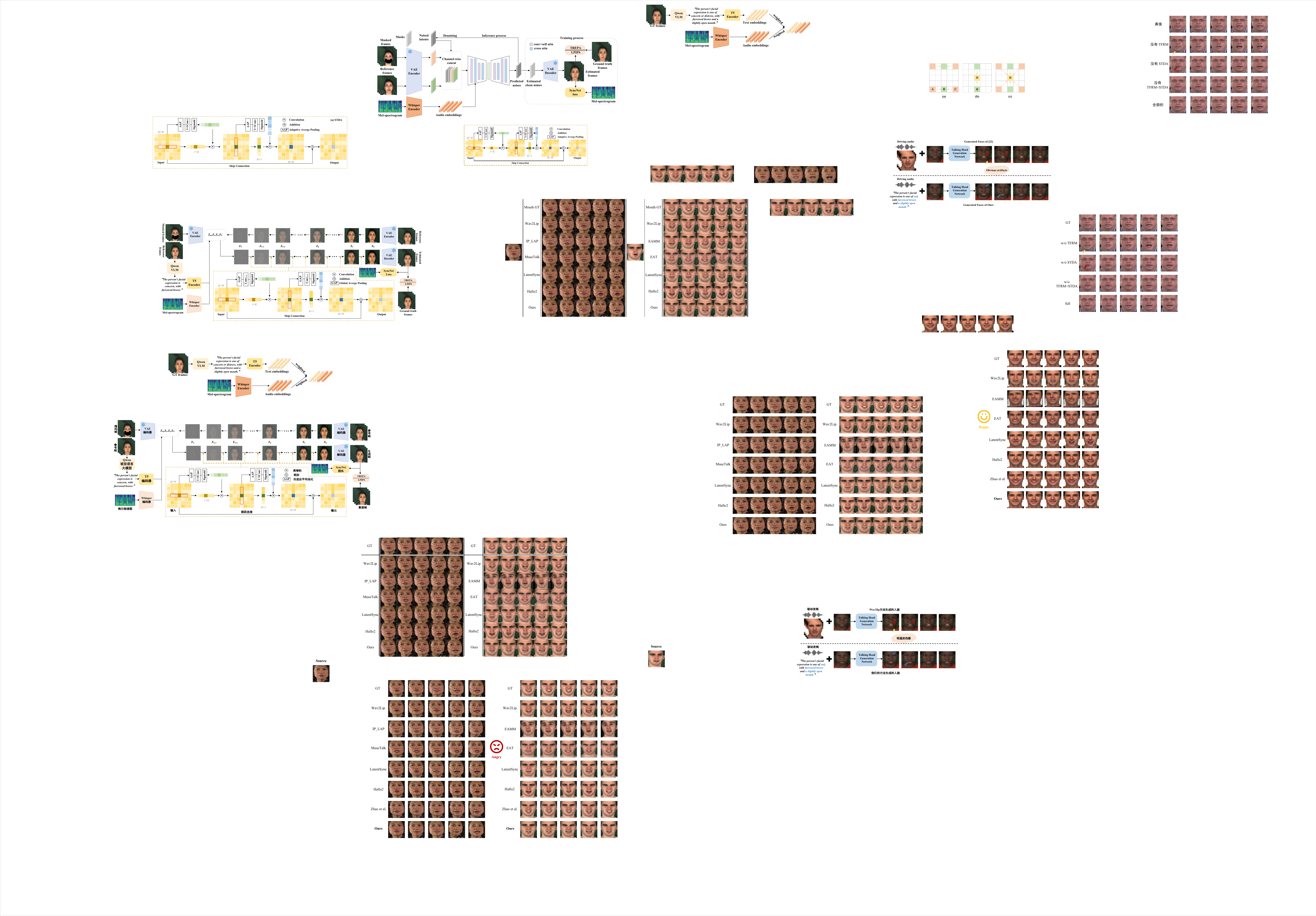}
    \caption{Audio-driven emotional generation synchronizes facial video with audio and aligns them with affective information. Previous methods suffer from visual artifacts (yellow arrows) and lip-sync issues (red arrows). Our method utilizes cross-modal information fusion to generate realistic faces that match the emotional text (blue arrows).}
    \label{fig:motivation}
\end{figure}

\section{Introduction}

In recent years, audio-driven talking head video generation technology has made significant progress, and its application potential in animation production, digital humans, and human-computer interaction \cite{bozkurt2023speculative, tian2024emo,fu2025fsdenet} is increasingly prominent. Previous methods \cite{cheng2022videoretalking, guan2023stylesync, stypulkowski2024diffused} focused on generating lip-sync talking head videos, neglecting the emotional changes that accompany human expression. Some recent methods \cite{ji2022eamm, gan2023efficient, ma2304talkclip, ma2023dreamtalk} have shifted the focus from lip-syncing to emotion control to generate emotionally rich talking heads. Early methods \cite{gan2023efficient, zhang2024emodiffhead} relied on discrete emotion labels to drive facial expressions in generated videos. While simple to implement, these methods struggle to capture the subtle dynamics of emotion. Other methods \cite{ji2022eamm, wang2023progressive} directly extract emotional information from emotional video templates. While these methods preserve more detail, they face the computational complexity of extracting dense emotional signals frame-by-frame. However, these methods struggle to capture the subtle dynamics and continuous changes in human emotions, resulting in limited emotional expressiveness in the generated results. Meanwhile, existing methods \cite{prajwal2020lip, shen2024decouple} also exhibit a clear trade-off between generation quality and dynamic conditional modeling. While Generative Adversarial Networks (GANs) can synthesize visually high-quality videos, they fall short in dynamic modeling of audio-visual synchronization and conditional dependencies, limiting their applicability in complex emotion generation scenarios.

Generative quality and emotional expression capability are two fundamental factors in improving the effectiveness of talking head video synthesis technology in practical applications~\cite{mattursun2024bss,zhang2024nonlinear, YinfengICLR2022saavn,yu2023measuring,li2025audio,zhang2025advancing,zhang2025iterative,yu2025dynamic}. However, existing research has rarely adequately addressed these two issues. To simultaneously address these two problems, this paper proposes an emotion-driven talking head generation framework based on a diffusion model. We employ a denoised U-Net as the backbone network, replacing traditional GANs with a diffusion model to enhance the conditional awareness of the generation process \cite{cao2024vnet}. To further improve temporal modeling and long-range dependency capture, we introduce a spatio-temporal directional attention mechanism into U-Net to enhance pixel-level long-range dependency modeling. Simultaneously, we design a temporal frame graph reasoning module to compensate for the shortcomings of the attention mechanism in inter-frame dependency modeling, achieving more coherent temporal generation. In terms of sentiment modeling, to overcome the computational burden of extracting sentiment signals frame by frame and the information loss of emotion labels, we propose a semantically guided sentiment condition construction method. First, we construct high-quality sentiment-text alignment data, then use a pre-trained QWen large language model \cite{bai2025qwen2} to extract sentiment descriptions from the alignment videos, and fuse the textual features with audio features to form fine-grained conditional inputs \cite{wang2025modality}. Finally, we generate sentiment-rich and temporally coherent talking head videos through a conditional diffusion process.

Overall, our main contributions are summarized as:

\begin{itemize}
\item We present a diffusion-based framework for generating emotional speech heads, achieving high-quality audio-visual synchronization and dynamic condition modeling through a denoised U-Net backbone network.
\item By introducing spatio-temporal directional attention module, our method enhances pixel-level long-range dependency modeling, further improving the accuracy of audio-visual synchronization.
\item We propose a temporal frame graph reasoning module that captures inter-frame temporal dependencies using a chain-structured graph, thereby improving temporal consistency.
\item We use large language model to extract fine-grained emotional semantic descriptions from aligned videos to achieve precise emotion control.
\end{itemize}

\section{Related Works}

\subsection{Audio-driven Talking Head Generation}

Deep learning-based audio-driven talking head generation methods \cite{prajwal2020lip, zhang2024musetalk, li2024latentsync, zhong2023identity, cui2024hallo2, mukhopadhyay2024diff2lip, shen2023difftalk, zhang2024emodiffhead} have gradually become a research hotspot. Early GAN-based methods \cite{prajwal2020lip, zhou2020makelttalk} can generate lip movements synchronized with speech, but often suffer from identity loss. To alleviate the identity preservation problem, IP\_LAP \cite{zhong2023identity} introduces multiple reference images as appearance priors, improving identity consistency. MuseTalk \cite{zhang2024musetalk} further achieves a breakthrough in inference efficiency, enabling real-time and identity-consistent head video generation. In terms of temporal consistency modeling, LatentSync \cite{li2024latentsync} proposes a temporal representation alignment method, which improves smoothness across frames without sacrificing lip movement synchronization accuracy. With the development of diffusion models, researchers have gradually introduced them into this field. Diff2Lip \cite{mukhopadhyay2024diff2lip} treats audio as a separate branch and incorporates reference images during generation to better preserve identity features. Hallo2 \cite{cui2024hallo2} can control facial expressions and postures by introducing adjustable text prompts, even supporting portrait animation synthesis lasting tens of minutes. The methods mentioned above mostly focus on architectural design, while another type of work focuses on the representation level, aiming to define and organize the characteristics themselves. Zhao et al. \cite{zhao2025synergizing} propose a multi-scale compensation codebook that decouples facial motion and identity appearance into independent spaces, enabling motion-guided appearance compensation via cross-modal queries. Building on these insights, this paper adopts a diffusion-based generation paradigm and introduces a temporal frame graph reasoning module to explicitly enforce motion smoothness across frames. In parallel, we further propose a spatio-temporal directional attention module that enhances pixel-level long-range dependency modeling.

\subsection{Audio-driven Emotion-aware Talking Head Generation}

In recent years, researchers have begun to incorporate emotional factors into audio-driven head-generating tasks. EVP \cite{ji2021audio} first proposes decoupling the emotion and content spaces from audio and mapping them onto facial keypoints, achieving emotionally expressive lip movement generation. Subsequently, EAMM \cite{ji2022eamm} uses emotional videos as the emotion source and enhances robustness through data augmentation, thereby generating emotionally expressive head animations under one-shot conditions. EAT \cite{gan2023efficient} achieves lightweight adaptation by introducing an emotional adaptation module and an emotional deformation network, enabling effective conversion from emotionless to emotionally expressive videos. As the advantages of diffusion models in generating quality have been validated, related methods have gradually incorporated them into emotion-driven generation. EmoTalker \cite{zhang2024emotalker}, based on a diffusion model, further models emotional intensity on top of audio-driven generation, thereby generating high-quality, emotionally expressive head videos. EMOdiffhead \cite{zhang2024emodiffhead} explores the specific representation of emotional conditions, combining facial expression vectors extracted from DECA \cite{feng2021learning} as emotional conditions to learn rich facial information from emotion-independent data. Simultaneously, it utilizes ReferenceNet to extract identity features from reference images to better preserve the person's identity during denoising. To introduce rich sentiment semantics while keeping the model lightweight, we propose using text generated by a large language model as a sentiment prior to guide the generation of the target speaker's sentiment expression in this paper.

{\begin{figure*}
   \centering
   \includegraphics[width=1\linewidth]{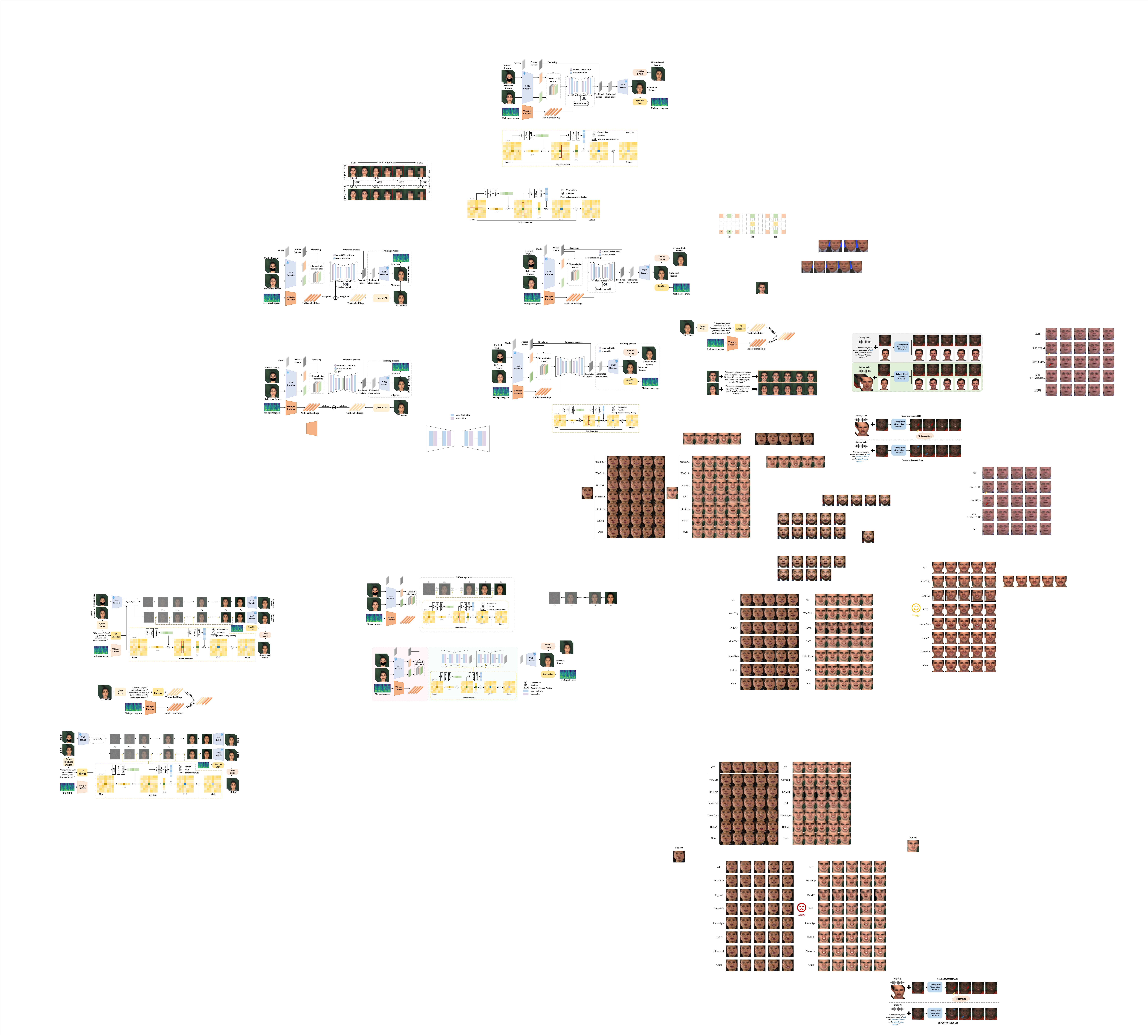}
    \caption{Overview of our framework. Audio features are extracted using Whisper, and emotional text is generated by Qwen-VLM and encoded with T5 encoder. These two modalities are fused into a multi-modal conditional embedding Z$_c$, which is then injected into the denoising U-Net. The U-Net incorporates a spatio-temporally dependent attention mechanism and a temporal graph reasoning module to model motion dynamics and ensure inter-frame consistency. The model is trained with TREPA, LPIPS and SyncNet loss functions to achieve accurate lip synchronization, rich emotional expression and natural dynamic effects. The latent noise Z$_\text{T}$, obtained after $\text{T}$ diffusion steps, serve as the starting point for the denoising process. The latent encodings of the masked input image and the reference identity image are represented as Z$_\text{m}$ and Z$_\text{r}$, respectively.}
    \label{fig:framework}
\end{figure*}}

\section{Methods}

\subsection{Overview}

As shown in Figure \ref{fig:framework}, we propose an emotion-aware video diffusion model. The model uses the U-Net architecture as the backbone denoising network and integrates three key components: Spatio-Temporal Directional Attention (STDA), Temporal Frame graph Reasoning Module (TFRM) and Text-audio fusion strategy and quality filtering. (1) STDA is integrated into the ResNet Block to model fine-grained correlations in low-level features, supplementing the receptive field limitation of convolutions and enhancing spatio-temporal perception. (Section 3.2); (2) TFRM constructs a frame-level graph neural network to explicitly model long-range cross-frame dependencies, ensuring the continuity and emotional consistency of the generated video in the temporal dimension (Section 3.3); (3) The text-audio fusion strategy based on quality filtering retains only training samples where the text description and audio emotion signal are aligned, ensuring that the fused multi-modal conditions are cleaner and more reliable for subsequent generation (Section 3.4). We combine all the losses in Section 3.5 to generate high-quality faces with lip-sync and emotion.

\subsection{Spatio-temporal directional attention mechanism}

Previous approaches \cite{shen2023difftalk, yin2022styleheat} employ a standard self-attention mechanism. However, the computational cost of this mechanism increases quadratically with the feature map size, easily leading to memory explosions when processing long sequences or multi-frame parallel training. Specifically, for self-attention, given an input tensor $\mathbf{X} \in \mathbb{R}^{C \times H \times W}$, self-attention can be formally expressed as:

\begin{align}
\mathrm{Attention}(\mathbf{Q},\mathbf{K},\mathbf{V})=\mathrm{Softmax}(\mathbf{Q}\mathbf{K}^{\top})\mathbf{V},
\end{align}

where $\mathbf{Q} = \mathbf{X}\mathbf{W}^Q$, $\mathbf{K} = \mathbf{X}\mathbf{W}^K$, $\mathbf{V} = \mathbf{X}\mathbf{W}^V$ are linear transformations of the input $\mathbf{X}$. As can be seen, the computational overhead of the self-attention mechanism mainly comes from: first, generating query ($\mathbf{Q}$), key ($\mathbf{K}$) and value ($\mathbf{V}$) tensors, with a computational complexity of $3HWC^2$; second, calculating the key-query dot product to generate the attention graph, with a complexity of $(HW)^2C$; and finally, weighted summation based on the attention graph, also with a complexity of $(HW)^2C$.

To reduce the overall complexity of these three steps and achieve efficient information aggregation, we propose the Spatio-Temporal Directional Attention mechanism (STDA), inspired by \cite{cui2024dual}, which consists of vertical and horizontal strip attention operators, as shown in the yellow dashed box in Figure \ref{fig:framework}. Since the operations in both directions share a similar pipeline, we only introduce the details of the horizontal pipeline here. Given an input feature $X \in \mathbb{R}^{C \times H \times W}$, instead of generating $\mathbf{Q}$, $\mathbf{K}$ and $\mathbf{V}$ like self-attention mechanisms, attention weights $\mathbf{A}$ are generated through a lightweight branch consisting of Global Average Pooling (GAP), 1×1 Convolution ($\mathrm{Conv}_{1\times1}$) and Sigmoid function ($\sigma$). The attention weights can be expressed as:

\begin{equation}
\mathbf{A}=\sigma(\mathrm{Conv}_{1\times1}(\mathrm{GAP}(X)))\in\mathbb{R}^{K},
\end{equation}

where $K$ specifies the length of a horizontal strip for integration. We share $\mathbf{A}$ across spatial and channel dimensions for efficiency.

The refined features are obtained through a convolution-style integration method with a computational complexity of $H \times W \times C \times K$. This approach is more efficient than self-attention, which requires $(H \times W)^2 \times C$ operations. The integration process can be formally expressed as:

\begin{equation}
\hat{\mathbf{X}}_{c,h,w}=\sum_{k=0}^{K-1}\mathbf{A}_{k}\mathbf{X}_{c,h,w-\lfloor\frac{K}{2}\rfloor+k},
\end{equation}

To summarize, the horizontal strip attention operator can be formally expressed as:

\begin{equation}
\mathbf{\hat{X}}=S_{K}^{\mathrm{H}}(\mathbf{X}),
\end{equation}

\begin{figure}[t]
    \centering
    \includegraphics[width=1\columnwidth]{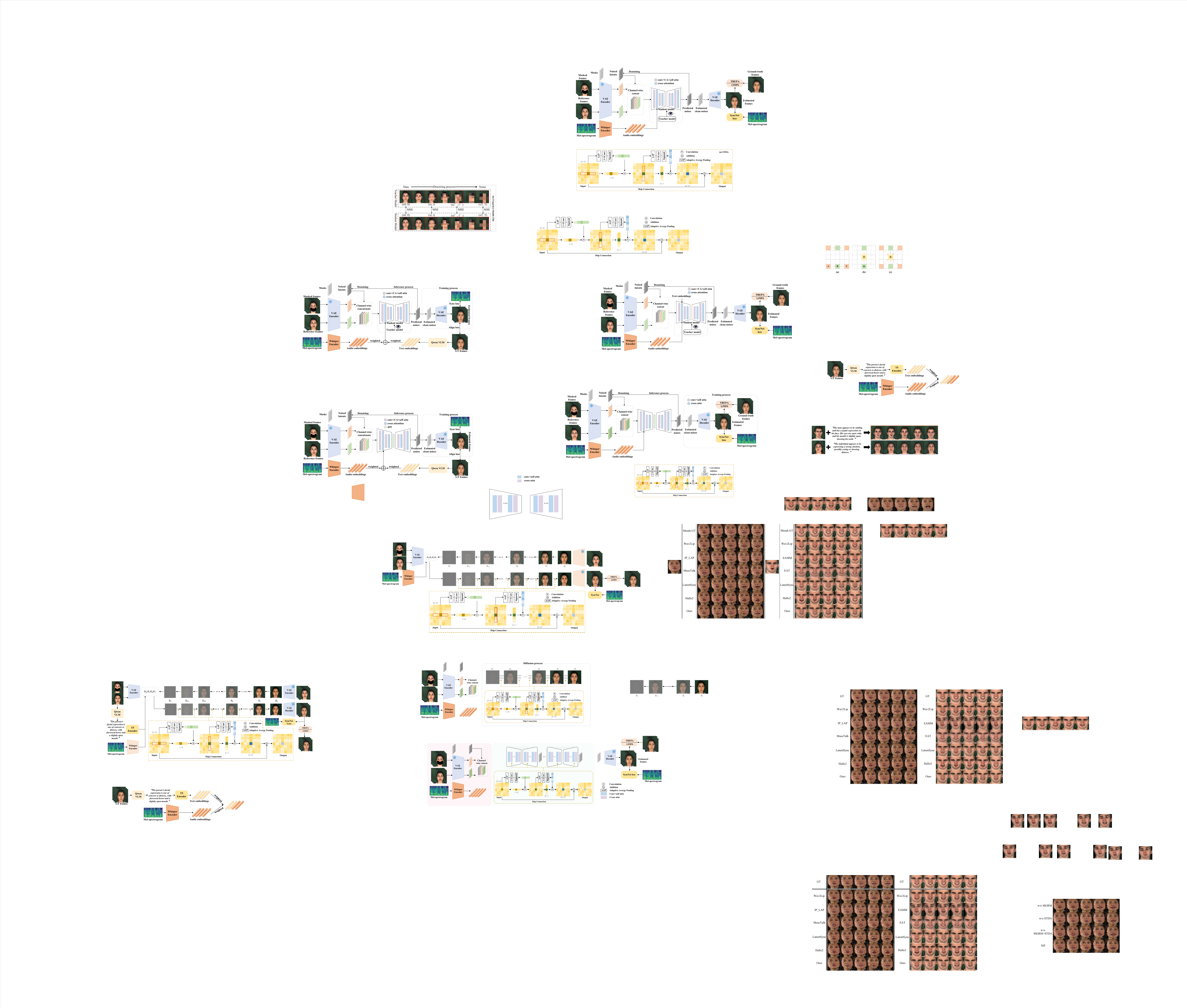}
    \caption{Different integration paradigms. (a) Horizontal strip attention. (b) Vertical strip attention. (c) Spatio-temporal directional attention (STDA) unit.}
    \label{fig:integration}
\end{figure}

The output of STDA is obtained by sequentially using horizontal and vertical strip attention operators as:

\begin{equation}
\mathbf{X}_{K}^{\mathrm{STDA}}=\mathrm{STDA}_{K}(\mathbf{X})=S_{K}^{\mathrm{V}}(S_{K}^{\mathrm{H}}(\mathbf{X})),
\end{equation}

where $S_K^{\mathrm{V}}(\cdot)$ denotes the vertical strip attention operators. By doing this, STDA achieves a receptive field comparable to global self-attention with linear complexity. We select only a few representative pixels for illustration, as shown in Figure \ref{fig:integration}, the horizontal and vertical attention operators integrate information in two directions, respectively. The horizontal one gives $\mathrm{B}=w_{AB}\mathrm{A}+w_{BB}\mathrm{B}+w_{CB}\mathrm{C}$, where $w_{ij}$ represents the attention weight from $i$ to $j$. Therefore, using $\mathrm{B}$ from Figure \ref{fig:integration}(a), $\mathrm{D}$ in Figure \ref{fig:integration}(b) can be represented as:

\begin{equation}
\mathrm{D}=w_{BD}\mathrm{B}+w_{DD}\mathrm{D},
\end{equation}

and then,

\begin{equation}
\mathrm{D}=w_{BD}(w_{AB}\mathrm{A}+w_{BB}\mathrm{B}+w_{CB}\mathrm{C})+w_{DD}\mathrm{D}.
\end{equation}

Based on this, the central pixel implicitly perceives the context of the entire region defined by $K \times K$.

\subsection{Temporal frame graph reasoning module}

Complex spatio-temporal dependencies exist between objects and actions in videos. Mainstream video diffusion models \cite{blattmann2023align, blattmann2023stable} typically employ temporal attention mechanisms to enhance inter-frame correlation modeling. However, these methods often model temporal dependencies as implicit attention weights, lacking explicit constraints on continuous motion trajectories between frames. This leads to issues such as motion jitter and identity drift in generated videos, especially with a significant decrease in temporal smoothness during long sequence generation. To address this problem, this paper proposes a Temporal Frame graph Reasoning Module (TFRM), which reformulates video temporal modeling as a graph structure learning task. This module explicitly models continuous dependencies between adjacent frames by constructing a chained temporal graph structure, thereby constraining the smoothness of motion trajectories and improving the temporal coherence.

\begin{algorithm}[t]
\caption{Temporal frame graph reasoning module}
\label{alg:TFRM}
\begin{algorithmic}[1]
\REQUIRE Latent features $\mathbf{X} \in \mathbb{R}^{B\times C\times T\times H\times W}$
\REQUIRE (Optional) object features $\mathbf{O} \in \mathbb{R}^{B\times T\times N\times C}$
\ENSURE Refined features $\mathbf{X}'$
\STATE Frame nodes: $\mathbf{F}_{b,t} = \text{AvgPool}_{h,w}(\mathbf{X}_{b,:,t,:,:})$
\STATE Temporal edges: $\mathcal{E}_f = \{(t,t{+}1),(t{+}1,t)\}$
\STATE Frame-level reasoning: $\mathbf{F}' = \text{GNN}_{f}(\mathbf{F}, \mathcal{E}_f)$
\STATE $\mathbf{X} \leftarrow \mathbf{X} + \text{Broadcast}(\mathbf{F}')$
\IF{object features $\mathbf{O}$ are available}
    \STATE Object reasoning: $\mathbf{O}' = \text{GNN}_{o}(\mathbf{O})$
    \STATE $\mathbf{G}_{b,t} = \frac{1}{N}\sum_{n}\mathbf{O}'_{b,t,n}$
    \STATE $\mathbf{X} \leftarrow \mathbf{X} + \text{Broadcast}(\mathbf{G})$
\ENDIF
\RETURN $\mathbf{X}$
\end{algorithmic}
\end{algorithm}

The TFRM is integrated into the core processing flow of UNetMidBlock, located between the cross-attention layer and the residual connection block. Specifically, as illustrated in Algorithm \ref{alg:TFRM}, each video frame feature $\mathbf{X} \in \mathbb{R}^{B\times C\times T\times H\times W}$ is compressed through spatial average pooling to obtain frame nodes $\mathbf{F}_{b,t} = \text{AvgPool}_{h,w}(\mathbf{X}_{b,:,t,:,:})$, resulting in a node feature tensor $\mathbf{F} \in \mathbb{R}^{B\times T\times C}$. From a graph perspective, this corresponds to constructing a node set $V = {v_1, v_2, ..., v_T}$, where each node $v_t$ represents the pooled feature of the $t$-th frame. For a video sequence containing $T$ frames, an edge set $\mathcal{E}_f = \{(t, t+1), (t+1, t) | t = 1, 2, ..., T-1\}$ is established, which ensures that each frame node only establishes bidirectional connections with its directly preceding and directly succeeding frames. Finally, the temporal graph, defined by node features $\mathbf{F}$ and edge set $\mathcal{E}_f$, is fed into a graph neural network $\text{GNN}_{f}$ for message passing, enabling the model to capture inter-frame dependency patterns and produce updated node representations. The updated features are reshaped into tensor dimensions of $[B, C, T, 1, 1]$, and added element-wise to the original latent features $[B, C, T, H, W]$ through residual connections. TFRM reserves an extension interface for object-level reasoning in its design. This branch theoretically requires input object features $\mathbf{O} \in \mathbb{R}^{B\times T\times N\times C}$. Generating such features requires running a pre-trained object detector and performing ROI Align operations on the spatial feature map $\mathbf{X}_{:,:,t,:,:} \in \mathbb{R}^{B\times C\times H\times W}$ for each frame, which significantly increases computational demands. Therefore, in the current implementation, we only use frame-level graph reasoning to model temporal dependencies.

\subsection{Text-audio fusion strategy and quality filtering}

{\bf Multi-modal condition fusion.} EVP \cite{ji2021audio} demonstrates that audio signals possess semantic ambiguity and interpretive differences, making it difficult to convey subtle emotional nuances through rhythm alone. To address this, we introduce sentiment-enriched textual descriptions as disambiguating priors. The audio stream provides precise lip synchronization and speech prosody, while the text stream resolves emotional ambiguity \cite{yu2025dope, yu2025dgfnet}. Our multi-modal fusion mechanism operates through a weighted feature integration strategy. Specifically, the textual features $f_{\mathrm{t}}(\mathbf{x})$ undergo linear transformation through projection parameters $W$ and $b$, followed by token-wise average pooling, before being scaled by a weighting factor $\lambda$. This is then added to the audio features of $\mathbf{x}$ to obtain the fused features $f_{\mathrm{c}}$. $f_{\mathrm{c}}(\mathbf{x})$ is subsequently injected into multiple layers of the U-Net encoder, enabling collaborative emotion control throughout the iterative denoising process. The above process can be represented as follows:

\begin{align}
f_{\mathrm{c}}(\mathbf{x}) &= f_{\mathrm{a}}(\mathbf{x}) + \lambda \cdot \mathrm{mean}\left(f_{\mathrm{t}}(\mathbf{x}) \cdot W^T + b\right),
\end{align}

where $\mathrm{mean}(\cdot)$ performs average pooling on the text tokens. $\lambda$ is a learnable weight, empirically initialized to 0.3.

{\bf Training set filtering.} To ensure cross-modal alignment and training stability, we filter the training set by retaining only samples with high cosine similarity between the fused feature and the ground-truth emotion embedding. Let $\mathbf{y}_{\mathbf{x}}$ denote the ground-truth emotion embedding of sample $\mathbf{x}$. The filtered dataset is defined as:

\begin{align}
\mathcal{D}_{\text{select}} &= \left\{ \mathbf{x} \in \mathcal{D}_{\text{total}} \mid \text{sim}\left(f_{\mathrm{c}}(\mathbf{x}), \mathbf{y}_{\mathbf{x}}\right) \geq \tau \right\}.
\end{align}

where $\mathrm{sim}(\cdot, \cdot)$ denotes cosine similarity and $\tau$ is a threshold empirically set to $0.8$ based on validation performance. This filtering removes samples with weak or misaligned cross-modal correspondence, preventing noisy examples from disturbing training.

\subsection{Training objective}

{\bf Noise prediction loss.} $\mathcal{L}_{\mathrm{noise}}$ constrains U-Net to predict a result as close as possible to the actual sampled noise $\epsilon$ based on the noisy latent variable $z_t$ and the conditional audio feature $\tau_{\theta}(A)$ at the current time step $t$:

\begin{equation}
\mathcal{L}_{\mathrm{noise}}=\mathbb{E}_{x,A,\epsilon\sim\mathcal{N}(0,1),t}\left[\left\|\epsilon-\epsilon_{\theta}(z_{t},t,\tau_{\theta}(A))\right\|_{2}^{2}\right],
\end{equation}

where $A$ is the input audio, $\tau_{\theta}$ is the audio feature extractor, and $\epsilon_\theta(z_t,t,\tau_\theta(A))$ is the predicted noise.

{\bf Lip sync Loss.} Our model predicts in the noise space, while SyncNet \cite{prajwal2020lip} requires input in the image space. To solve this, we reconstruct a one-step estimate of the clean latent $\hat{z}_0$ from the current noisy latent $z_t$ using the DDPM \cite{ho2020denoising} inversion formula:

\begin{equation}
\hat{z}_0=\begin{pmatrix}z_t-\sqrt{1-\bar{\alpha}_t}\epsilon_\theta(z_t)\end{pmatrix}/\sqrt{\bar{\alpha}_t},
\end{equation}

where $\bar{\alpha}_t$ is the cumulative noise schedule. This one-step approximation allows us to efficiently obtain a plausible clean latent without running the full reverse diffusion process.

The estimated $\hat{z}0$ is then decoded into an RGB image via the VAE decoder $\mathcal{D}(\cdot)$. To enhance the discriminative power of the synchronization loss, we extend the input of SyncNet from its default single-frame-pair setting to 16 consecutive frames. Specifically, for a sliding window starting at frame index $f$, we feed SyncNet with the decoded video clip $\mathcal{D}(\hat{z}_0)_{f:f+16}$ and its corresponding audio segment $a_{f:f+16}$. The synchronization loss is defined as:

\begin{equation}
\mathcal{L}_{\mathrm{sync}}=\mathbb{E}_{x,a,\epsilon,t}\left[\mathrm{SyncNet}(\mathcal{D}(\hat{z}_0)_{f:f+16},a_{f:f+16})\right],
\end{equation}

where $\mathbb{E}$ is over random time steps $t$ and frame windows $f$.

{\bf LPIPS Loss.} We also use LPIPS \cite{zhang2018unreasonable} to supervise the visual details of the images generated by U-Net.

\begin{equation}
\mathcal{L}_{\mathrm{lpips}}=\mathbb{E}_{x,\epsilon,t}\left[\left\|\mathcal{V}_l(\mathcal{D}(\hat{z}_0)_f)-\mathcal{V}_l(x_f)\right\|_2^2\right],
\end{equation}

where $\mathcal{V}_{l}(\cdot)$ represents the features extracted from the $l^{th}$ layer of the pre-trained VGG network.

{\bf TREPA Loss.} $\mathcal{L}_{\mathrm{trepa}}$ enhances temporal consistency by aligning generated and real sequence representations.

\begin{equation}
\mathcal{L}_{\mathrm{trepa}}=\mathbb{E}_{x,\epsilon,t}\left[\left\|\mathcal{T}(\mathcal{D}(\hat{z}_0)_{f:f+16})-\mathcal{T}(x_{f:f+16})\right\|_2^2\right],
\end{equation}

where $\mathcal{T}$ is the encoder of the self-supervised video model \cite{wang2023videomae}.

The total loss is calculated as follows:

\begin{equation}
\mathcal{L}_{\mathrm{total}}=\lambda_1\mathcal{L}_{\mathrm{noise}}+\lambda_2\mathcal{L}_{\mathrm{sync}}+\lambda_3\mathcal{L}_{\mathrm{lpips}}+\lambda_4\mathcal{L}_{\mathrm{trepa}}.
\end{equation}

Through empirical analysis, we determine the optimal values for the coefficients $\lambda_1$, $\lambda_2$, $\lambda_3$ and $\lambda_4$ to be 1, 0.05, 0.1 and 10, respectively.

\section{Experiment}

\subsection{Datasets and Metrics}

{\bf Datasets.} We use HDTF \cite{zhang2021flow} and MEAD \cite{wang2020mead} datasets as our training sets. HDTF contains 362 different high-definition videos, typically ranging in resolution from 720p to 1080p. To evaluate sentiment accuracy on the basis of the speaking head generation task, we also train our model using the MEAD dataset, a corpus of speaking face videos containing 60 actors speaking with eight emotions at three different intensities. We use HyperIQA \cite{su2020blindly} to filter out videos with low visual quality, particularly blurry or pixelated ones. To verify the generalization ability of our model, we also design a cross-dataset generalization experiment using RAVDESS. RAVDESS \cite{livingstone2018ryerson} contains videos of 24 completely different actors in 8 different emotions, 2 different intensities, and only frontal viewpoints. During evaluation, we randomly select 30 videos from the test sets of both datasets.

{\bf Metrics.} We evaluate our method in four aspects: (1) Visual quality. We use SSIM and FID \cite{heusel2017gans} to evaluate visual quality. (2) Lip-sync accuracy. We use the confidence score of SyncNet (Sync$_{conf}$) \cite{chung2016out}. (3) Temporal consistency. We adopt the widely used FVD \cite{unterthiner2018towards}. (4) Sentiment accuracy. Following EAT \cite{gan2023efficient}, we adopt a two-level attention-based framework to compute sentiment accuracy (Acc$_{emo}$). A Frame Attention Network (FAN) extracts frame-wise features and generates attention weights, which are aggregated into video-level representations for softmax classification across eight emotion categories. Since HDTF lacks sentiment annotations, we report Acc$_{emo}$ only on MEAD.

{\begin{figure*}
   \centering
   \includegraphics[width=0.9\linewidth]{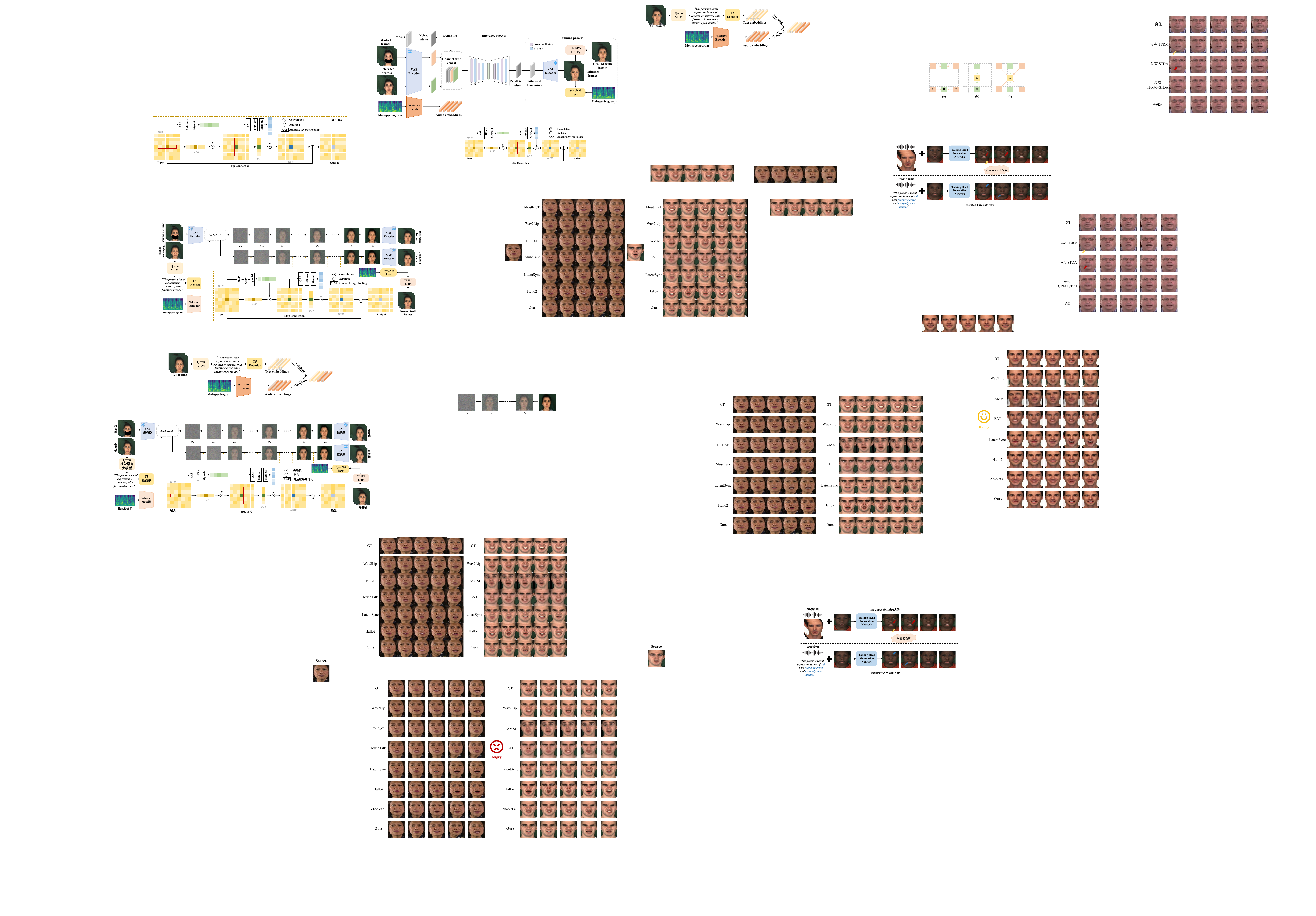}
    \caption{Qualitative comparison with SOTA methods. The left generated faces are from the HDTF dataset. The right generated faces are from the MEAD dataset, demonstrating the "angry" emotion.}
    \label{fig:view}
\end{figure*}}

\subsection{Implementation Details}

We use scenedetect to detect scene transition points in each video. Based on the detected scene transition points, the original video is segmented into multiple small segments, and ffmpeg is used to cut the video into 5-second segments. An affine transformation is performed on the faces in each frame to center and normalize them. All video sequences are uniformly downsampled to 25 FPS, and an affine transformation is performed on the facial feature points detected by the face alignment algorithm to extract 256×256 face video segments. Audio is resampled to 16 kHz. During inference, a 20-step DDIM sampling method is used to generate the final result.

\subsection{Results and Discussions}

{\bf Quantitative comparisons.} Table \ref{tab:hdtf_results} and Table \ref{tab:mead_results} illustrate the advantages of EAD-Net. Our method achieves the highest lip-sync accuracy on both datasets, likely due to the supervision of SyncNet and the audio cross-attention mechanism, which may better capture the relationship between
audio and lip movements. Regarding visual quality, Wav2Lip performs best on the SSIM metric through its local editing strategy, which we analyze as preserving the original facial structure outside the mouth region. Our method outperforms other methods on the FVD metric on both datasets. On the MEAD dataset, we also note that our method achieves the best FID value. We believe this may be attributed to STDA, through its directional strip attention mechanism, aggregates spatial context information in the horizontal and vertical directions, enhancing the fine-grained texture realism and structural coherence of the generated frames. In terms of sentiment accuracy, our method achieves the best emotional accuracy, demonstrating the advantages of text-audio fusion strategies in high-level semantic control.

{\bf Qualitative comparisons.} Figure \ref{fig:view} presents the qualitative comparisons. Wav2Lip generates artifacts in the mouth and chin regions on both datasets. On the HDTF dataset, IP\_LAP shows little variation in lip shape. The lip region generated by MuseTalk, especially the teeth, exhibits blurring. On the MEAD dataset, EAMM shows significant distortion and blurring artifacts in the eye region, with its mouth remaining essentially unchanged across five consecutive frames. EAT performs slightly better in lip shape fidelity, but its eye perspective is inconsistent with ground truth. On both datasets, LatentSync maintains reasonable overall structure, but its lip texture is blurry and lacks detail. Hallo2 has texture enhancement issues with high facil contrast, likely due to enhancing high-frequency signals for high-resolution (4K) generation. Furthermore, it shows gaze direction deviations in consecutive frames. Zhao et al.'s method has visual realism but still exhibits lip shape discrepancies. Our method generates realistic lip shapes on both datasets, accurately aligning with the audio while preserving identity features.

\begin{figure}[t]
    \centering
    \includegraphics[width=0.9\columnwidth]{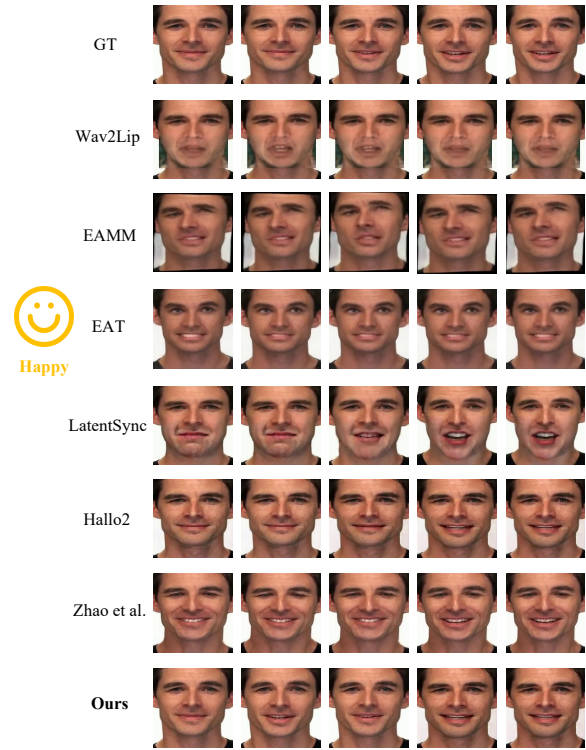}
    \caption{Visualization of cross-dataset generalization results. The model trained on MEAD is directly applied to RAVDESS to generate "happy" expressions.}
    \label{fig:ravdess}
\end{figure}

{\bf Generalization evaluation.} Figure \ref{fig:ravdess} shows a visual comparison of various methods on the cross-dataset test set. Wav2Lip exhibits identity bias, where the generated lip regions shift towards the speaker characteristics of the training set. EAMM still has local deformations in the eye region and the lip movement amplitude is small. Although EAT explores the diversity of head poses, deviations from the target viewpoint are observed. LatentSync exhibits localized appearance degradation around the mouth, particularly noticeable shadowing in the lip region. The methods of Hallo2 and Zhao et al. perform well in terms of visual realism, with lip shapes close to real videos, but the method of Zhao et al. has blurry teeth details, and the accuracy of lip contours still has slight deviations, and the eye details are insufficient. In contrast, although our method still exhibits slight deviations from the real video in terms of lip shape, the rhythm of lip movements remains well synchronized with the audio. Identity features are well preserved, and no obvious artifacts are observed.

\begin{table}[t]
\centering
\caption{Quantitative comparisons on HDTF dataset. ↑ indicates the higher the better, while ↓ indicates the lower the better. We denote the best scores in bold and second-best \underline{underlined}.}
\label{tab:hdtf_results}
\begin{tabular}{l|cccc}
\toprule
Method & FID↓ & SSIM↑ & Sync$_{conf}$↑ & FVD↓ \\
\midrule
Wav2Lip \cite{prajwal2020lip} & 32.57 & {\bf 0.8751} & 8.03 & 459.543 \\
IP\_LAP \cite{zhong2023identity} & 31.41 & 0.8180 & 2.33 & 404.340 \\
MuseTalk \cite{zhang2024musetalk} & 27.91 & 0.8316 & 4.20 & 353.756 \\
LatentSync \cite{li2024latentsync} & \underline{26.28} & \underline{0.8675} & \underline{8.91} & 332.257 \\
Hallo2 \cite{cui2024hallo2} & 53.47 & 0.6668 & 5.82 & 413.951 \\
Zhao et al. \cite{zhao2025synergizing} & {\bf 23.27} & 0.7947 & 5.88 & \underline{315.236} \\
\textbf{EAD-Net (Ours)} & 29.91 & 0.8639 & {\bf 9.04}  & {\bf 312.347} \\
\bottomrule
\end{tabular}
\end{table}

\begin{table}[t]
\centering
\caption{Quantitative comparisons on MEAD dataset. ↑ indicates the higher the better, while ↓ indicates the lower the better. We denote the best scores in bold and second-best \underline{underlined}.}
\label{tab:mead_results}
\resizebox{\linewidth}{!}{
\begin{tabular}{l|ccccc}
\toprule
Method & FID↓ & SSIM↑ & Sync$_{conf}$↑ & FVD↓ & Acc$_{emo}$↑ \\
\midrule
Wav2Lip \cite{prajwal2020lip} & 45.62 & {\bf 0.8855} & 6.65 & 321.913 & 37.14\% \\
EAMM \cite{ji2022eamm} & 156.15 & 0.4427 & 2.03 & 661.170 & 23.33\% \\
EAT \cite{gan2023efficient} & 87.56 & 0.6673 & 6.12 & 705.216 & 35.10\% \\
LatentSync \cite{li2024latentsync} & 44.12 & 0.8544 & \underline{7.41} & 295.258 & 43.33\% \\
Hallo2 \cite{cui2024hallo2} & 42.41 & 0.6808 & 5.08 & 346.330 & 44.45\% \\
Zhao et al. \cite{zhao2025synergizing} & \underline{42.05} & 0.8382 & 6.11 & \underline{274.013} & \underline{45.11\%} \\
\textbf{EAD-Net (Ours)} & {\bf 41.81} & \underline{0.8576} & {\bf 7.45}  & {\bf 265.082} & {\bf 46.67\%} \\
\bottomrule
\end{tabular}
}
\end{table}

\subsection{Ablation Study}

\begin{table}[t]
\centering
\caption{Ablation study of components on MEAD dataset.}
\label{tab:ablation_mead_quantity}
\resizebox{\columnwidth}{!}{
\begin{tabular}{ccc ccccc}
\toprule
\multicolumn{3}{c}{\textbf{Components}} & \multicolumn{5}{c}{\textbf{Metrics}} \\
\cmidrule(r){1-3} \cmidrule(l){4-8}
STDA & TFRM & Text & FID↓ & SSIM↑ & Sync$_{conf}$↑ & FVD↓ & Acc$_{emo}$↑ \\
\midrule
\ding{55} & \ding{55} & \ding{55} & 44.12 & 0.8544 & 7.41 & 295.258 & 43.33\% \\
\ding{51} & \ding{55} & \ding{55} & 45.89 & 0.8547 & {\bf 7.79}  & 285.459 & 43.19\% \\
\ding{55} & \ding{51} & \ding{55} & 42.67 & 0.8550 & 7.50  & {\bf 262.353} & 43.33\% \\
\ding{55} & \ding{55} & \ding{51} & 43.50 & 0.8547 & 7.42  & 285.974 & 46.10\% \\
\ding{51} & \ding{51} & \ding{55} & 52.20 & 0.8557 & 7.47 & 284.655 & 43.33\% \\
\ding{55} & \ding{51} & \ding{51} & \underline{42.32} & \underline{0.8577} & \underline{7.70} & 301.730 & 45.23\% \\
\ding{51} & \ding{55} & \ding{51} & 45.30 & {\bf 0.8590} & 7.14 & 307.599 & \underline{46.51\%} \\
\ding{51} & \ding{51} & \ding{51} & {\bf 41.81} & 0.8576 & 7.45 & \underline{265.082} & {\bf 46.67\%} \\
\bottomrule
\end{tabular}
}
\end{table}

In this section, we conduct ablation experiments on the MEAD and HDTF datasets to validate the effectiveness of the core components of our method. Quantitative results are shown in Table \ref{tab:ablation_mead_quantity} and Table \ref{tab:ablation_hdtf_quantity}. Furthermore, we present qualitative ablation results on the HDTF dataset. STDA, TFRM and Text represent our core components. The symbol \ding{51} indicates the addition of the module, while \ding{55} indicates the absence of the module.

\begin{table}[t]
\centering
\caption{Ablation study of components on HDTF dataset.}
\label{tab:ablation_hdtf_quantity}
\resizebox{0.8\columnwidth}{!}{
\begin{tabular}{cc cccc}
\toprule
\multicolumn{2}{c}{\textbf{Components}} & \multicolumn{4}{c}{\textbf{Metrics}} \\
\cmidrule(r){1-2} \cmidrule(l){3-6}
STDA & TFRM & FID↓ & SSIM↑ & Sync$_{conf}$↑ & FVD↓ \\
\midrule
\ding{55} & \ding{55} & {\bf 26.28} & 0.8675 & 8.91 & 332.257 \\
\ding{51} & \ding{55} & 30.80 & 0.8636 & \underline{9.00}  & 319.795 \\
\ding{55} & \ding{51} & \underline{28.56} & {\bf 0.8678} & 8.85 & {\bf 281.089} \\
\ding{51} & \ding{51} & 29.91 & \underline{0.8639} & {\bf 9.04} & \underline{312.347} \\
\bottomrule
\end{tabular}
}
\end{table}

\begin{figure}[t]
    \centering
    \includegraphics[width=0.9\columnwidth]{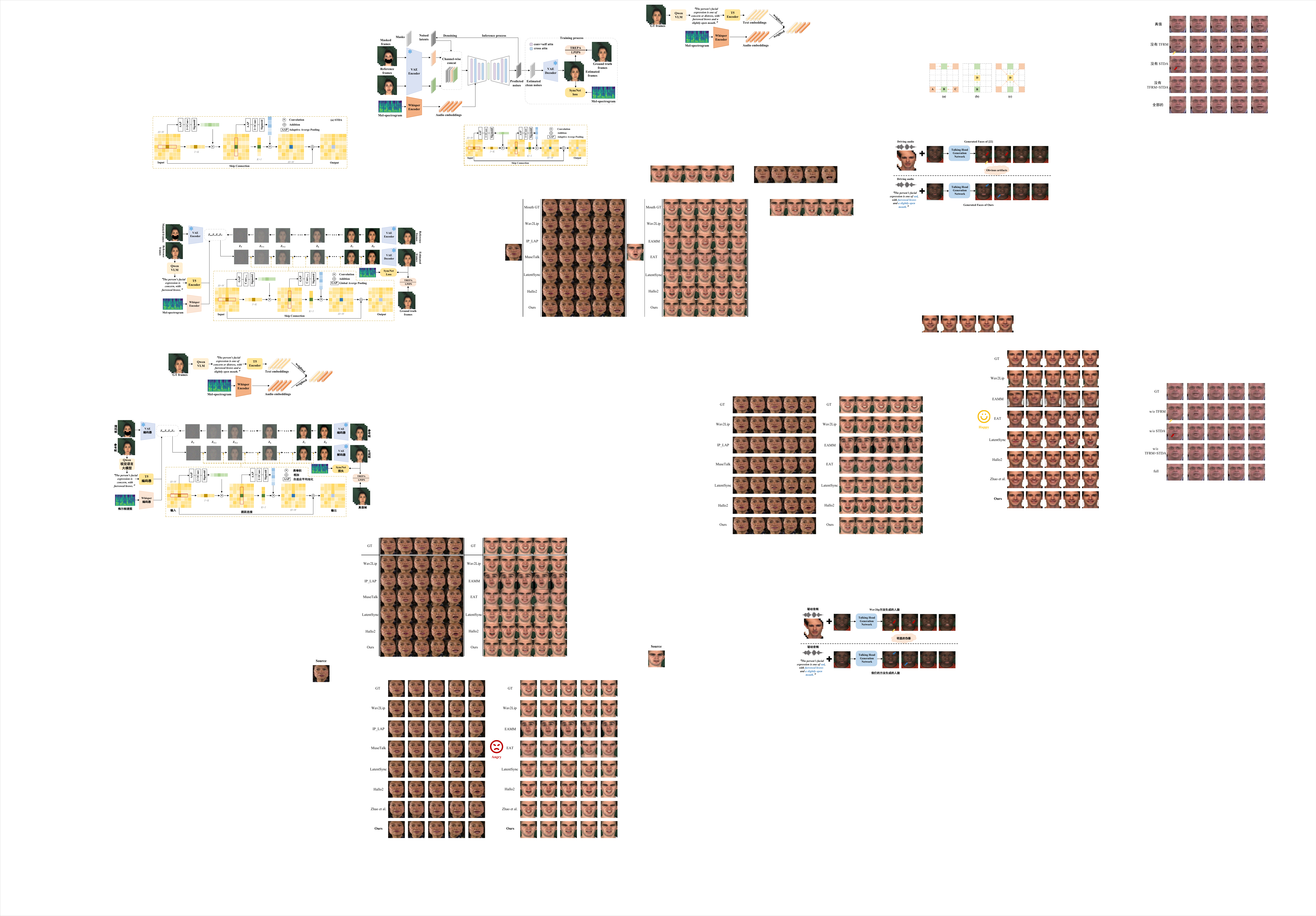}
    \caption{Qualitative results of the ablation experiments performed on model components on the HDTF dataset.}
    \label{fig:ablation_hdtf}
\end{figure}

{\bf Ablation experiments on the MEAD dataset with different components.} Table \ref{tab:ablation_mead_quantity} shows that introducing STDA alone improves Sync$_{conf}$ to a best value of 7.79 and reduces FVD from 295.258 to 285.459, validating the central role of the spatio-temporal attention mechanism in audio-visual alignment and temporal dependency modeling. Introducing TFRM alone achieves the best FVD value (262.353), demonstrating its effectiveness in enhancing temporal consistency. Introducing Text alone improves Acc$_{emo}$ to 46.10\%, confirming that text generated by large language models can serve as a lightweight yet effective semantic prior for controlling sentiment expression. When STDA and TFRM are combined, FID increases to 52.20 and Sync$_{conf}$ drops to 7.47. We believe this is because TFRM suppresses high-frequency inter-frame variations, which STDA primarily utilizes to establish fine-grained pixel correspondences. This representational incompatibility leads to degraded fidelity and lip synchronization. After introducing the semantic prior from Text (i.e., when all three modules work together), the model achieves the lowest FID (41.81), the highest Acc$_{emo}$ (46.67\%), and a near-optimal FVD (265.082). The global facial semantic layout provides an alternative pathway for STDA, allowing it to integrate semantic priors with mid-to-low-frequency features. This reduces STDA‘s reliance on high-frequency information and bypasses the bottleneck caused by TFRM. STDA and Text together achieve the highest SSIM (0.8590) and the second-highest Acc$_{emo}$ value (46.51\%), demonstrating strong complementarity between audio-visual alignment and high-level semantic conditions. The combination of TFRM and Text achieves the second-lowest FID (42.32) while maintaining competitive SSIM (0.8577) and Sync$_{conf}$ (7.70), demonstrating that this combination improves visual fidelity and lip-sync on the MEAD dataset. We note that adding Text slightly increases FVD compared with the baseline, which stems from semantically guided motion variations that introduce additional temporal dynamics despite improved expressiveness.

{\bf Ablation experiments on the HDTF dataset with different components.} Table \ref{tab:ablation_hdtf_quantity} shows that introducing STDA alone can improve Sync$_{conf}$ to 9.00, and reduce FVD to 319.795, verifying the positive impact of the spatio-temporal attention mechanism on audio-visual alignment and temporal coherence. However, FID increases from 26.28 to 30.80, and SSIM decreases to 0.8636, indicating that the spatio-temporal constraints introduced by STDA reduce single-frame fidelity to some extent. Using TFRM alone yields the best FVD (281.089), demonstrating its advantage in modeling long-range motion dependencies. However, Sync$_{conf}$ drops to 8.85, indicating that TFRM's focus on temporal smoothness comes at a slight cost to lip-sync precision. STDA and TFRM working together achieve the best lip synchronization, proving that STDA's audio-visual alignment capability and TFRM's temporal modeling capability can effectively complement each other. However, FVD increases from 281.089 (TFRM alone) to 312.347, indicating a slight trade-off between the fine-grained alignment of STDA and the global smoothness enforced by TFRM. Figure \ref{fig:ablation_hdtf} visually illustrates the impact of each component. With STDA introduced, the lip shape is similar to the ground truth, but noticeable artifacts exist in the neck region (the yellow arrows). When TFRM is used alone, the facial movement transitions are natural, however, the matching degree between the lips and the ground truth decreases slightly (the red arrows). Integrating both components achieves improved alignment in the lip region. While this results in a slight decrease in temporal coherence compared to using TFRM alone, it still significantly outperforms the baseline model.

\subsection{Limitation and Future Work}

While our approach has made progress, several aspects still require improvement. Currently, the model only supports eight predefined emotion categories with fixed intensities, lacking continuous control over emotion intensity. Furthermore, current research primarily focuses on frontal viewpoint generation, while multi-view datasets like MEAD offer possibilities for head pose control. Inspired by EmoTalkingGaussian \cite{cha2025emotalkinggaussian}, Future work will explore continuous emotion modeling using valence and arousal as conditions, transcending the eight-category limit to represent arbitrary emotion types and intensities. Second, we will investigate end-to-end head pose generation without intermediate representations, drawing on GaussianHeads \cite{teotia2024gaussianheads} to treat head pose as a learnable parameter jointly optimized with facial geometry using multi-view video, enhancing pose diversity and realism.

\section{Conclusion}

This study proposes an emotion-aware diffusion framework called EAD-Net for generating talking head videos with accurate lip-sync, rich emotional expression, and temporal consistency. Our method utilizes a large language model to extract semantic text and combines it with audio features to constrain the U-Net. To address the lip-sync challenge in video generation, this framework integrates the SyncNet supervision with TREPA, effectively improving audio-visual alignment accuracy. Simultaneously, we design a spatio-temporal directional attention to capture spatio-temporal correlations across frames and spaces through pixel-level long-range dependency modeling. Additionally, a temporal frame graph reasoning module is introduced to model inter-frame temporal dependencies. Experiments show that our framework performs well in lip sync, timing consistency and emotional expression.

\section{Acknowledgments}

This study was funded by the National Natural Science Foundation of China (grant numbers 62463029, 62472368, and 62303259).

\bibliographystyle{ACM-Reference-Format}
\bibliography{main}

\end{document}